\documentclass[runningheads]{lncs} 

\usepackage{graphics} 
\usepackage{epsfig} 
\usepackage{mathptmx} 
\usepackage{times} 
\usepackage{amsmath} 
\usepackage{amssymb}  
\usepackage{comment}
\usepackage{xspace}
\usepackage{xcolor}

\newcommand{\dynsys}{\ensuremath{\mathcal{D}}\xspace}

\newcommand{\nats}{\ensuremath{\mathbb{N}}\xspace}
\newcommand{\reals}{\ensuremath{\mathbb{R}}\xspace}

\newcommand{\ftraj}{\ensuremath{\mathbf{F}}\xspace}
\newcommand{\btraj}{\ensuremath{\mathbf{B}}\xspace}
\newcommand{\nextt}{\ensuremath{\mathit{next}}\xspace}
\newcommand{\prev}{\ensuremath{\mathit{prev}}\xspace}
\newcommand{\noise}{\ensuremath{E}\xspace}
\newcommand{\model}{\ensuremath{\mathcal{M}}\xspace}

\newcommand{\init}{\ensuremath{I}\xspace}
\newcommand{\trans}{\ensuremath{F}\xspace}
\newcommand{\ctrl}{\ensuremath{u}\xspace}
\newcommand{\timestep}{\ensuremath{\delta}\xspace}
\newcommand{\horizon}{\ensuremath{T}\xspace}
\newcommand{\reach}{\ensuremath{G}\xspace}
\newcommand{\avoid}{\ensuremath{A}\xspace}
\newcommand{\state}{\ensuremath{\vec{x}}\xspace}

\newcommand{\stateset}{\ensuremath{\mathcal{X}}\xspace}
\newcommand{\State}{\stateset}

\title{\LARGE \bf
A New Strategy for Verifying Reach-Avoid Specifications in Neural Feedback Systems
}

\author{Samuel I. Akinwande\inst{1}, Sydney M. Katz\inst{1}, Mykel J. Kochenderfer\inst{1}, and Clark Barrett\inst{2}
}
\institute{
Department of Aeronautics and Astronautics, Stanford University, Stanford, USA \and
Department of Computer Science, Stanford University, Stanford, USA
}

\authorrunning{Akinwande et al.}

\begin{document}
\maketitle

\begin{abstract}
Forward reachability analysis is the predominant approach for verifying reach-avoid properties in neural feedback systems—dynamical systems controlled by neural networks. This dominance stems from the limited scalability of existing backward reachability methods. In this work, we introduce new algorithms that compute both over- and under-approximations of backward reachable sets for such systems. We further integrate these backward algorithms with established forward analysis techniques to yield a unified verification framework for neural feedback systems.
\end{abstract}

\section{Introduction}
Neural feedback systems—dynamical systems controlled by neural networks—are becoming increasingly prevalent, with applications in robotics, autonomous driving \cite{ettinger2021large}, aerospace autonomy \cite{kaufmann2023champion}, and beyond. As these systems proliferate, the need to ensure compliance with safety and design specifications has become increasingly critical. In particular, it is essential to verify that neural feedback systems operate as intended under all relevant conditions.

Common approaches for verifying specification satisfaction include sampling, invariance analysis, and reachability analysis \cite{akinwande2025verifying,delecki2025diffusion,mandal2024formally}. 
Sampling, a form of falsification, seeks to identify violations of a desired specification by evaluating a finite set of scenarios. 
This approach is widely used in industrial practice due to its scalability and generality, but it cannot provide formal guarantees. 
Invariance analysis, by contrast, offers formal guarantees by synthesizing a certificate function—typically a Lyapunov and/or barrier function—that certifies specification satisfaction. 
However, such certificates are often difficult to construct in practice. 
Reachability analysis evaluates all possible system states to determine whether a given property is satisfied or violated. 
It provides formal guarantees and is applicable to a broad class of systems, but suffers from a trade-off between precision and scalability: precise methods scale poorly due to the curse of dimensionality.

Most existing methods for reachability analysis of neural feedback systems rely on forward analysis, propagating the system dynamics forward to compute all possible states. 
This reliance arises because computing a backward pass through a neural network is particularly challenging \cite{rober2023backward}. 
However, the exclusive use of forward analysis often amplifies the inherent limitations of reachability analysis. 

In this work, we explore a strategy that combines an existing forward analysis method with new techniques for backward analysis of neural feedback systems, aiming to mitigate some of these limitations.

\section{Problem Statement}
Following \cite{akinwande2025verifying}, we define a discrete-time neural feedback system $\dynsys$ as the tuple
\[
\langle n, \init, \trans, \noise, \ctrl, \timestep, \horizon, \reach, \avoid \rangle,
\]
where $n \in \nats$ is the \emph{dimension} of the system (i.e., each state is an element of $\reals^n$), 
$\init \subseteq \reals^n$ is the set of \emph{initial states}, 
$\trans = (f_1, \dots, f_n)$ is a sequence of \emph{state update functions} with $f_i : \reals^n \to \reals$, 
$\noise \subseteq \reals^n$ is the \emph{perturbation set} (i.e., the set of admissible disturbance terms that may affect the next state), 
$\ctrl : \reals^n \to \reals^n$ is the \emph{control function}, 
$\timestep \in \reals_+$ is the \emph{time-step size}, 
$\horizon \in \nats$ is the \emph{number of time steps}, 
$\reach \subseteq \reals^n$ is the set of \emph{goal states}, 
and $\avoid \subseteq \reals^n$ is the set of \emph{avoid states} (i.e., unsafe states).

States evolve over a sequence of $\horizon$ discrete time steps, each of duration $\timestep$, giving a total time horizon of $\timestep \cdot \horizon$. 
For a current state $\state \in \reals^n$, the next-state function $\nextt^{\dynsys} : \reals^n \to 2^{\reals^n}$ defines the possible successor states. 
This set-valued mapping captures the nondeterminism introduced by the perturbation term. 
For each $i \in [1, \ldots, n]$,
\begin{align}
    \nextt^{\dynsys}(\state)_i 
    = \big\{ \state_i + \big(f_i(\state) + \ctrl(\state)_i + \epsilon_i\big) \cdot \timestep \;\big|\; \epsilon \in \noise \big\},
    \label{eq:fwd_diffEq}
\end{align}
Note that $f_i$ and $\ctrl$ may depend on the entire previous state $\state$, not only on the component $\state_i$. 
We assume that the neural network controller $\ctrl$ is a fully-connected, feed-forward network with $n$ inputs, $n$ outputs, and ReLU activations.

The previous-state function $\prev^{\dynsys} : 2^{\reals^n} \to 2^{\reals^n}$ defines the \emph{preimage} (or backward projection) of the forward evolution operator, and is given by
\begin{align}
    \prev^{\dynsys}(\State)
    = \big\{\, \vec{y} \in \reals^n \;\big|\; \nextt^{\dynsys}(\vec{y}) \subseteq \State \,\big\}.
    \label{eq:bw_diffEq}
\end{align}
We define a forward trajectory $\ftraj^{\dynsys}(\State_0)$, as the sequence of state sets $(\State_0, \State_1, \dots, \State_T)$, where $\State_i = \nextt^{\dynsys}(\State_{i-1})$ for all $i \in [1, T]$. 
Similarly, we define a backward trajectory $\btraj^{\dynsys}(\State_T)$, where $\State_T$ is a target set, as the sequence of state sets $(\State_T, \State_{T-1}, \dots, \State_0)$, where $\State_i = \prev^{\dynsys}(\State_{i+1})$ for all $i \in [0, T-1]$.

We say that a system $\dynsys$ is \emph{safe} if it satisfies either the forward or backward reach-avoid properties. 
The \emph{forward reach-avoid properties} are given by Equations~\eqref{eq:fwd_props}:
\begin{subequations}
\begin{align}
    &\forall\, \state_0 \in \init.\: \exists\, t \in [0..T].\: \ftraj^{\dynsys}(\{\state_0\})_t \subseteq \reach, \label{eq:fwd_reach_prop}\\
    &\forall\, t \in [0..T].\: \ftraj^{\dynsys}(\init)_t \cap \avoid = \emptyset, \label{eq:fwd_avoid_prop}
\end{align}
\label{eq:fwd_props}
\end{subequations}
and the \emph{backward reach-avoid properties} are given by Equations~\eqref{eq:bw_props}:
\begin{subequations}
\begin{align}
    &\exists\, t \in [0..T].\: \btraj^{\dynsys}(\reach)_t \supseteq \init, \label{eq:bw_reach_prop}\\
    &\forall\, t \in [0..T].\: \btraj^{\dynsys}(\avoid)_t \cap \init = \emptyset. \label{eq:bw_avoid_prop}
\end{align}
\label{eq:bw_props}
\end{subequations}

Equations~\ref{eq:fwd_reach_prop} and~\ref{eq:bw_reach_prop} are the \emph{reach} properties. 
They express that every forward trajectory starting from a state in the initial set eventually reaches the goal set, or that every backward trajectory starting from the terminal set traces back to the initial set within the specified number of time steps. 
Conversely, Equations~\ref{eq:fwd_avoid_prop} and~\ref{eq:bw_avoid_prop} are the \emph{avoid} properties, which require that system trajectories avoid unsafe states at every time step within the given horizon. 

Although the forward and backward properties are similar in intent, they differ subtly in interpretation. 
Forward properties assert that all trajectories initiated from the initial set satisfy the reach-avoid conditions, whereas backward properties characterize the set of initial states from which such safe trajectories can originate. 

Formally, we aim to establish that $\dynsys$ is safe. 
Equations~\ref{eq:fwd_props} and~\ref{eq:bw_props} are \emph{sufficient} to verify safety, but relying solely on either forward or backward reachability analysis is limited by the challenges discussed earlier.
Our goal is to develop a verification strategy that jointly leverages Properties~\ref{eq:fwd_props} and~\ref{eq:bw_props} to soundly and effectively establish system safety.

\section{Proposed Methods}
\subsection{Forward Reachability Analysis}
Forward reachability analysis verifies forward reach-avoid properties by solving Equation~\ref{eq:fwd_diffEq} over $\horizon$ time steps for all initial states in $\init$. 
We employ the method of \cite{akinwande2025verifying} to compute an overapproximation of the true forward reachable set.

\subsection{Backward Reachability Analysis}
To compute backward reachable sets, we must solve Equation~\ref{eq:bw_diffEq}. 
The key challenge is that no scalable algorithms currently exist for directly solving Equation~\ref{eq:bw_diffEq}. 
Our objective, therefore, is to develop algorithms that can soundly approximate this equation in a computationally tractable manner. 
Given a system $\mathcal{D}$ and a target set $\mathcal{T}$, we can compute a lower bound on Equation~\ref{eq:bw_diffEq} by solving the following problem.
\begin{subequations}
\begin{align}
    \min \; \state_i^t, \\
    \text{subject to } 
    \quad \state_i^{t+1} = \state_i^t + \big(f_i(\state^t) + \ctrl(\state^t)_i + \epsilon_i\big) \cdot \timestep, \\
    \quad \model_0, \dots, \model_n, \quad
    \epsilon \in \noise, \quad
    \state_i^{t+1} \in \mathcal{T}.
\end{align}
\label{eq:bw_Reach_outer}
\end{subequations}

\noindent
Replacing the minimization with a maximization yields an upper bound. 
Together, these bounds define a hyperrectangle that provides a guaranteed overapproximation of the true backwards reachable set (BRS). 
We exploit this property to compute an under approximation of the BRS by formulating a constrained optimization problem: identify the largest subset of the overapproximation whose overapproximated forward image remains within the target set. 
We outline three methods for solving this problem.

\subsubsection{Golden Section Search (GSS)}
This method estimates the largest admissible under hyperrectangle via a golden section search. The overapproximate hyperrectangle has center $\vec{c}$ and radii $\vec{r}$, and the underapproximate hyperrectangle is parameterized as $\vec{c} \pm \rho \vec{r}$ with $\rho \in (0,1)$. We seek the maximal $\rho$ such that the overapproximated forward image of the under hyperrectangle remains within the target set.
\subsubsection{Iterative Convex Hull (ICH)}
Because GSS may require many forward queries, we propose a sampling-based alternative. We densely sample the overapproximate hyperrectangle, propagate each point using an inexpensive point query, and label samples as positive or negative depending on containment in the target set. The positive samples define a \emph{candidate} underapproximate hyperrectangle, which is validated using an overapproximated query. If containment fails, the candidate becomes the new overapproximation, and the process repeats.
\subsubsection{Largest Empty Box (LEB)}
For highly nonconvex reachable sets, we instead solve a small MILP to find the largest \emph{candidate} underapproximate hyperrectangle that excludes all negative samples from the ICH procedure. This candidate is then evaluated and refined using the same containment check.

\subsection{The FaBRe Verification Strategy}
Rather than exclusively computing forward or backward reachable sets, we propose verifying $\dynsys$ by combining both analyses. 
Specifically, we partition the planning horizon into $\horizon = F + B$ steps, where $F$ denotes the number of forward-analysis steps and $B$ the number of backward-analysis steps. 
The subdivision is treated as a configurable parameter of the solver. This yields the \textbf{F}orward \textbf{a}nd \textbf{B}ackward \textbf{Re}achability analysis strategy.

To establish that the system satisfies a reach property (Properties~\ref{eq:fwd_reach_prop} and~\ref{eq:bw_reach_prop}), we perform $F$ steps of forward analysis from $\init$ and $B$ steps of backward analysis from $\reach$. 
If the overapproximation of the forward reachable set at time step $F$ is contained within the underapproximation of the backward reachable set at time step $B$, then the system is guaranteed to reach the goal set. 

To verify that the system satisfies an avoid property at time $t$ (Properties~\ref{eq:fwd_avoid_prop} and~\ref{eq:bw_avoid_prop}), we compute $F$ steps of overapproximated forward analysis from the initial set and $B$ steps of overapproximated backward analysis from the avoid set $\avoid$. 
If the resulting forward and backward reachable sets are disjoint for all $t \in [1, T]$, then the system is guaranteed to avoid all unsafe states, thereby ensuring safety.

\subsection{Ongoing Work}
We are currently evaluating the performance of our proposed approach relative to existing state-of-the-art methods. 
In particular, we are comparing our overapproximation method with that of \cite{rober2023backward}, our under-approximation method with \cite{sidrane2025burns}, and the FaBRe strategy with the state-of-the-art forward reachability method introduced in \cite{akinwande2025verifying}. 
Our goal is to demonstrate measurable improvements in scalability over current approaches.

\newpage
\bibliographystyle{plain}
\bibliography{refs}
\end{document}